\documentclass[manuscript, screen, nonacm]{acmart}

\usepackage{graphicx}
\usepackage{amsmath}
\usepackage{booktabs}

\title{Aligned Machine: A Benchmark for Evaluating Multimodal Embedding Similarity via Human Comparative Judgements}

\title[Aligned Machine]{Aligned Machine: A Benchmark for Evaluating Multimodal Embedding Similarity via Human Comparative Judgements}

\begin{document}

\author{Jiechen Li}
\email{jiechen.li@duke.edu}
\author{Hannah Groos}
\email{hannah.groos@duke.edu}
\author{Brinnae Bent}
\email{brinnae.bent@duke.edu}
\affiliation{%
  \institution{Duke University}
  \city{Durham}
  \state{NC}
  \country{USA}
 }









\begin{abstract}
Embedding models increasingly inform how people search, recommend, and retrieve information, yet we lack credible ways to test whether these similarity judgments reflect how humans understand meaning. Existing benchmarks typically rely on static, expert-labeled pairs with coarse Likert-style annotations, limiting scalability and making it difficult to obtain consistent, high-confidence annotations. We introduce Aligned Machine, an interactive, public-facing platform that transforms broad participation into a continuously growing Human-Aligned Similarity Benchmark (HASB) for evaluating semantic alignment in embedding-based systems. Participants complete a cognitively-natural, forced-choice task, comparing two pairs of items (text or images) and selecting which pair is more similar. This comparative design is accessible to non-experts, yields clean preference data, and supports robust aggregation. We report results from an initial deployment with 115 participants contributing 2,875 total labeled pairs and 872 unique pairs. Using HASB, we evaluated 12 widely used embedding models and found only partial alignment between embedding similarity and human judgments. Text-only models consistently achieve higher overall agreement than multimodal models, while multimodal performance varies systematically by task type: alignment is strongest for image–image comparisons, weaker for image–text comparisons, and lowest for text–text comparisons. Aligned Machine is designed to evolve over time as participation grows, supporting longitudinal analysis, domain-specific studies, bias detection, and measurement of concept drift across model versions. By combining accessible public participation with rigorous comparative evaluation, HASB provides a scalable and accountable framework for auditing semantic alignment in embedding models.
\end{abstract}

\maketitle

\section{Introduction}

Embedding models play a central role in decision-making, as their representations of semantic similarity drive retrieval-augmented generation (RAG), AI search, and recommendation systems. Rather than directly determining downstream decisions, embeddings serve as foundational components in retrieval and ranking pipelines, where similarity estimation shapes system behavior. Yet the similarity scores produced by these models often do not align with human intuition \cite{Vulic2020MultiSimLex}. As a result, misalignment at this representational layer exposes alignment gaps in systems that rely on embeddings.

Prior studies have shown that similarity relations learned by embedding models often differ from how humans judge meaning, especially across languages, domains, and cultural contexts \cite{Vulic2020MultiSimLex, aroyo2015truth}. This mismatch reflects broader concerns that embedding-based representations often prioritize statistical regularities in training data over semantic distinctions grounded in human interpretation \cite{bender2021stochastic, faruqui2016problems}. There exists a gap in current evaluation practice: performance on end-user tasks or static similarity datasets does not reveal whether embedding representations align with human semantic judgments.

Despite growing recognition that embedding representations often diverge from human semantic understanding, existing evaluation practices provide limited ways to examine this misalignment directly. Existing datasets such as STS-B \cite{cer2017semeval} and SimLex-999 \cite{hill2015simlex} rely on static, often expert-labeled pairs with coarse Likert-style annotations. These approaches are limited in scope and vulnerable to inter-annotator inconsistencies \cite{agirre2015semeval}. Moreover, these benchmarks typically rely on small groups of expert annotators, implicitly treating expert consensus as a proxy for human semantic understanding. This design limits the diversity of perspectives captured in evaluation and obscures how non-expert users interpret meaning across social, cultural, and contextual differences. As a result, dominant or expert-aligned interpretations are often overrepresented, while alternative or minority understandings remain largely invisible. These constraints make it difficult to examine whether embedding models capture the semantic distinctions that matter in public-facing contexts, where judgments are inherently comparative and situated.

To address this gap, we introduce the Aligned Machine, an interactive, public-facing platform that turns broad participation into a benchmark for evaluating semantic alignment in embedding-based systems. Inspired by participatory studies such as Moral Machine \cite{awad2018moral}, the platform invites people to complete simple comparative tasks in which they are shown two pairs of items, either text–text (TT), image–image (II) or mixed image–text (IT) pairs, and asked to select which pair appears more semantically similar (e.g., “Is (A, B) more similar than (C, D)?”). Using a binary, forced-choice format, our platform captures comparative judgments that are intuitive for non-experts and yield structured preference data for evaluation. These judgments allow model-induced similarity relations to be compared directly with aggregated human judgments, making alignment observable at the level of embedding representations. Collected at scale, these comparisons form the basis of a continuously evolving dataset, the Human-Aligned Similarity Benchmark (HASB), designed to support rigorous model evaluation alongside meaningful public engagement. HASB v1 introduces an evaluation paradigm intended to expand methodological approaches to embedding assessment, serving as an initial release of this framework. By focusing on similarity relations rather than task-specific outcomes, HASB enables alignment to be examined directly within embedding representations, without relying on application performance or assuming a single, fixed notion of semantic correctness.

In this work, we reframe semantic alignment in embedding models as a crowdsourced, comparative evaluation problem that can be examined directly at the representational level. This initial release focuses on lexical-level and cross-modal comparison in a controlled setting, with sentence- and document-level extensions forming part of future iterations. We use HASB to compare widely used text and multimodal embedding models, showing how their similarity judgments differ from human judgments.

\section{Relevant Work}

\subsection{Embedding-based Semantic Similarity}
Modern embedding models map linguistic and multimodal content into vectors, where semantic relatedness is measured using distance metrics \cite{mikolov2013efficient, pennington2014glove, devlin2019bert, radford2021clip}. To evaluate how well these representations capture human intent of similarity, researchers have largely relied on benchmark datasets built from human judgments \cite{hadjtaieb2020survey}.

Despite their importance in practice, embedding models are still commonly evaluated using static datasets such as STS-B \cite{cer2017semeval}, SimLex-999 \cite{hill2015simlex}, the SemEval-2012 Task \cite{jurgens2012semeval}, and word–sentence semantic relatedness dataset \cite{glasgow2016evaluating}. Prior work has identified several limitations of these datasets. First, coarse Likert-style annotations often fail to capture subtle differences in human judgment \cite{wadhwa2023rescale}. Second, expert-driven curation limits coverage and diversity \cite{raji2021everything}. Third, substantial disagreement among annotators persists even after aggregation \cite{kiela2020hateful}. More fundamentally, these datasets evaluate isolated item pairs, which makes it difficult to capture relational similarity structures. Prior work has noted that models optimized for these benchmarks may not generalize reliably beyond the evaluation setting, as commonly used automatic metrics correlate weakly with human judgments \cite{stiennon2020summarize}.

\subsection{Human-Centered Evaluation of AI Systems}
Prior work across NLP, cognitive science, and AI ethics has consistently shown that human judgments are subjective, context-dependent, and shaped by individual and cultural differences \cite{tversky1977features, medin1993respects, aroyo2015truth, pavlick2019inherent}. In research on human judgment, disagreement is often interpreted not as annotation error, but as evidence that multiple interpretations can be reasonable \cite{snow2008cheap}. This perspective is also supported by empirical work. For example, large-scale annotation efforts in natural language inference and visual recognition demonstrated that even simple decision tasks can produce systematic and meaningful variation in public judgments at scale \cite{nie2020adversarial, peterson2019uncertainty}.

Recent crowdsourcing approaches partially address these issues. The Moral Machine paradigm introduced scalable pairwise choices (A vs. B), which have since been adapted for large language model alignment \cite{takemoto2024moral, vida2024multilingual}. However, these approaches still focus on single items rather than relational comparisons. Psychologically, binary forced-choice comparisons between item pairs significantly reduce cognitive load and enhance annotation reliability compared to absolute rating scales like Likert \cite{kiritchenko2017bws}. However, these approaches have not been systematically applied to evaluating semantic similarity at the level of embedding representations, particularly for relational comparisons between single item pairs.

\subsection{Cognitive Foundations of Comparative Similarity}

Comparative similarity judgments have long been studied in psychophysics and cognitive science. Classical comparative judgment theory models pairwise comparison as a latent decision process along a shared psychological continuum \cite{thurstone1927law}. Subsequent extensions distinguish impersonal judgment from personal preference while maintaining a common comparative mechanism \cite{andrich2019law}. Cognitive research further shows that similarity judgments rely on structural alignment processes that prioritize relational correspondences over surface features \cite{markman2000structure}, and that similarity judgments reflect distinct relational computations rather than simple difference detection \cite{ichien2023asymmetry}.

Neuroimaging evidence suggests that conceptual representations activated by pictures and words exhibit overlapping structure \cite{dirani2023crossmodal}, and crossmodal accounts position similarity along a perceptual–conceptual continuum rather than as strictly modality-specific processing \cite{distefano2024perceptual}. These findings provide theoretical grounding for modeling semantic comparison behavior across modalities. In particular, the comparative nature of similarity judgment suggests that evaluating relative similarity between candidate pairs reflects a cognitively natural decision process rather than an artificial annotation task. This perspective motivates the use of pair-of-pairs comparative judgments as a structured way to capture how humans organize semantic relationships across both linguistic and visual representations.

\section{Methods}

\subsection{Aligned Machine System Design}
The Aligned Machine system implements a controlled experimental pipeline for collecting human comparative similarity judgments. 

\subsubsection{Concept Curation}
Textual concepts were derived from the SimLex-999 dataset \cite{hill2015simlex}. All unique words were extracted to form a vocabulary pool, and the original SimLex similarity scores were discarded. Words were randomly shuffled using a fixed random seed to ensure reproducibility and grouped into disjoint base pairs. Each word appears in only one base pair, ensuring that lexical concepts are not reused across pairs. This procedure preserves lexical diversity while preventing leakage of expert-provided similarity judgments into benchmark construction. Comparison instances are subsequently formed by pairing two base pairs, producing the pair-of-pairs structure used in the human evaluation task.

To enable multimodal evaluation, each word was mapped to a representative image. Image mappings were constructed using a staged sourcing strategy. Words were first matched to images from MS-COCO 2014 \cite{lin2014coco, coco_dataset}, followed by Flickr30k Dataset \cite{young2014flickr30k, flickr30k_dataset} for remaining terms. For words not covered by these datasets, images were manually selected from Unsplash \cite{unsplash}. Word-to-image mappings were stored independently of word pairs, enabling flexible recombination across modalities.

\subsubsection{Trial Construction and Modality Balancing}
At the core of the system is a pre-generated comparison space consisting of ordered pairs of semantic item pairs. Each comparison instance is defined as a tuple
\[
C = \langle (x_1, x_2), (y_1, y_2), m \rangle
\]
where $(x_1, x_2)$ and $(y_1, y_2)$ are candidate item pairs and $m \in \{\mathrm{TT}, \mathrm{IT}, \mathrm{II}\}$ denotes the presentation modality.

All comparison trials are generated offline prior to data collection. For each base word pair, all valid modality instantiations are constructed (text–text, image–text in both directions, and image–image). This exhaustive generation produces a large candidate pool from which session-specific trials are sampled. 

The candidate pool contains 100 TT comparisons (11.5\%), 100 II comparisons (11.5\%), and 672 IT comparisons (77.1\%). The concept sets are stored separately by modality and released on Hugging Face [HuggingFace dataset link redacted for double blind submission].

This skew toward IT comparisons reflects the substantially larger combinatorial space of cross-modal pairings. While TT and II comparisons are drawn from relatively constrained sets, IT comparisons involve combinations between text concepts and image concepts, resulting in many distinct pairings. As a result, when comparisons are randomly sampled and presented to users, IT comparisons are more likely to be unique across sessions. Although these comparisons differ at the instance level, they are grouped under the same IT modality during analysis, leading to a higher aggregate count for this category.

Sampling is performed under two constraints: (1) no word pair appears more than once within a session, and (2) modality distribution within a session is approximately balanced across TT, IT, and II conditions.

Each participant session consists of 25 trials drawn from this pool. Trials are randomly ordered within the session. Because the full comparison space is generated deterministically prior to sampling, all trials are reproducible and auditable, and the same underlying comparison units can be reused across participants.

Each participant session is assigned a unique anonymized identifier used to associate all trial responses within that session. The system records, for each trial, the comparison identifier, modality type, participant choice, and response timing. No personally identifiable information is stored. This structure supports session-level aggregation, modality-stratified analysis, and reproducible reconstruction of the experimental conditions under which each judgment was collected.

The system interface can be viewed at [redacted for double-blind submission]. 

\subsection{Data Collection}
Human similarity judgments were collected through a web-based application deployed over a three-month period. The study protocol was reviewed and approved by the [redacted for double blind submission] University Institutional Review Board (IRB \#2026-0014). All participants provided informed consent prior to participation. Participants (N=115) aged 18+ were recruited via public posts on social media. 

Each participant completed a single session consisting of 25 comparative similarity judgments. In each trial, participants were presented with two candidate pairs and asked to select the pair that appeared more semantically similar. Trials were randomized within each session. No feedback was provided during task completion. After completing all trials, participants were shown a descriptive summary of their responses; this information was presented after data collection and did not influence subsequent judgments. To preserve participant privacy, no personally identifiable information was stored. User identifiers were anonymized using randomly generated UUIDs, and the system recorded only comparative choices and associated metadata (e.g., modality type and response timing).

\subsection{Benchmark Data Curation}
Raw participant responses were filtered to remove incomplete, duplicated, or invalid sessions. After cleaning, responses from 115 participants were retained. These responses were converted into a frozen v1 benchmark dataset consisting of 919 TT comparisons (32\%), 1039 II comparisons (36.1\%), and 917 IT comparisons (31.9\%). The v1 benchmark is released as a fixed dataset to ensure that reported evaluation results remain comparable over time and can be found here: [HuggingFace dataset link redacted for double blind submission].

\subsection{Model Evaluation on HASB}
Embedding models were evaluated by comparing their similarity-based preferences against human judgments in HASB.

\subsubsection{Model Selection}
The selection of embedding models for evaluation in this study was guided by two primary criteria: empirical performance as indicated by the Massive Text Embedding Benchmark (MTEB) ranking on Hugging Face \cite{muennighoff2023mteb, mteb_huggingface}, and recency or prominence of release by major research organizations and technology companies. The final set comprises twelve models, including seven text embedding models and five multimodal models (Table 1). 

For ease of presentation and readability in subsequent analyses and visualizations, we adopt abbreviated model names throughout the paper. Table 1 summarizes the full model names, corresponding abbreviations, and model types used in this study.

\begin{table}
\caption{Embedding models evaluated in this study, including full model names, abbreviated labels used throughout the paper, and model types.}
\label{tab:freq}
\begin{tabular}{clp{0.62\linewidth}l}
\toprule
\# & Abbreviation & Model Name & Model Type \\
\midrule
1  & GEM-E   & gemini-embedding-001 & Text \\
2  & QW3-06  & Qwen/Qwen3-Embedding-0.6B & Text \\
3  & E5-LI   & intfloat/multilingual-e5-large-instruct & Text \\
4  & TE3L    & text-embedding-3-large & Text \\
5  & ML6     & sentence-transformers/all-MiniLM-L6-v2 & Text \\
6  & BGE-B   & BAAI/bge-base-en-v1.5 & Text \\
7  & GEMM-3  & google/embeddinggemma-300m & Text \\
8  & CLIP-B32 & openai/clip-vit-base-patch32 & Multimodal \\
9  & CLIP-B16 & openai/clip-vit-base-patch16 & Multimodal \\
10 & CLIP-L14 & openai/clip-vit-large-patch14 & Multimodal \\
11 & SIG-B16  & google/siglip-base-patch16-224 & Multimodal \\
12 & CLIP-H14 & laion/CLIP-ViT-H-14-laion2B-s32B-b79K & Multimodal \\
\bottomrule
\end{tabular}
\end{table}

\subsubsection{Embedding Models and Similarity Computation}
The evaluation includes both text-only and multimodal embedding models. Text-only models were evaluated exclusively on TT comparisons, while multimodal models were evaluated on TT, IT, and II comparisons. All models were integrated through a unified interface that standardizes preprocessing, embedding extraction, and similarity computation.

For each model, cosine similarity \cite{salton1983introduction} was computed between embeddings for each candidate pair. For II comparisons, multimodal models were evaluated using their image encoder towers. During inference, these image encoders produce image embeddings independently of the text encoder, effectively functioning as image-only embedding models for the purpose of computing similarity scores.

Model-specific normalization procedures were applied where required. Although raw cosine similarities can be negative, multimodal models such as SIG-B16 are trained with a sigmoid loss that prioritizes positive similarity signals. For presentation, we normalize the similarity scores to a positive range using L2 normalization. Differences in similarity output ranges across models were preserved rather than rescaled.

\subsubsection{Definition of Model–Human Agreement}
To assess alignment between human similarity judgments and embedding model behavior, we employ a four-step preference-matching procedure at the comparison level. Each comparison $i$ consists of a binary choice between two candidate pairs, Option~A and Option~B, as presented to participants.

First, for each comparison $i$, we retrieve the participant’s recorded choice $H_i \in \{A, B\}$.

Second, for each embedding model $m$, we obtain similarity scores for both options from a precomputed similarity table, denoted as $s^{m}_{A}$ and $s^{m}_{B}$.

Third, we induce the model’s discrete preference by comparing these two scores. The induced model preference $P^{m}_{i}$ is defined as:
\[
P^{m}_{i} = \arg\max_{o \in \{A,B\}} s^{m}_{i,o}.
\]

Finally, we define per-comparison model--human agreement as an indicator of whether the induced model preference matches the human choice:
\[
a^{m}_{i} = \mathbb{I}\!\left[P^{m}_{i} = H_i\right],
\]
where $\mathbb{I}[\cdot]$ represents the indicator function.

Per-comparison agreement indicators $a^{m}_{i}$ serve as the basic units for computing alignment accuracy. Aggregation and normalization across modalities are described in Section~3.4.4.

\subsubsection{Preference-Based Alignment Metric}
Model performance was assessed using a preference-matching metric. For each human comparison between two candidate pairs (A and B), a model’s predicted preference was determined by comparing its similarity scores for the two options. A prediction was considered correct if the model’s preferred option (highest similarity score) matched the human choice.

To ensure fair comparison across modalities with differing numbers of trials, alignment accuracy was computed separately for each modality and then macro-averaged. Let $\mathcal{K} = \{\mathrm{TT}, \mathrm{IT}, \mathrm{II}\}$. Overall alignment accuracy is defined as:
\[
\mathrm{Acc} = \frac{1}{|\mathcal{K}|} \sum_{k \in \mathcal{K}} \mathrm{Acc}_k,
\]
where $\mathrm{Acc}_k$ denotes the alignment accuracy computed for modality $k$.

When a modality contains multiple datasets, alignment accuracy is first macro-averaged within the modality (with equal weight per dataset) before averaging across modalities.

To quantify uncertainty in the reported alignment accuracies, we compute 95\% confidence intervals for each model. Error bars represent these intervals, estimated using the t-distribution based on the variability of agreement scores across all evaluated pairs for each model.

\subsubsection{Inter-Annotator Variability}
To characterize human judgment variability, we conduct analyses at two complementary levels, examining between-pair variability and within-pair variability.

In the between-pair analysis, each unique item pair is treated as the unit of analysis. For each pair, we compute the proportion of annotations selecting option B, allowing us to assess how human preferences vary across different semantic comparisons.

To assess inter-annotator agreement on the same semantic comparison, we further analyze within-pair variability for pairs that received multiple annotations. For each pair, we compute within-pair variance using the binomial variance formulation
\[
\mathrm{Var}(X) = p(1 - p)
\]
where $p$ represents the proportion of annotators selecting option~B. This variance ranges from $0$ (perfect agreement, where all annotators choose the same option) to a maximum of $0.25$ (maximum disagreement, where exactly half choose A and half choose B; $\mathrm{Var}_{\max} = 0.5(1 - 0.5)$), providing a normalized measure of disagreement that is comparable across pairs with different annotation counts.

Item pairs are randomly generated and independently sampled across users, such that annotation counts vary due to random exposure rather than inherent controversy of the pairs. This design allows within-pair variance to be interpreted as reflecting genuine diversity in human semantic judgments rather than a sampling bias.

In addition, we examine the relationship between annotation count and within-pair variance to assess how observed disagreement changes with annotation depth. For each multi-annotated pair, we compute the Pearson correlation between the number of annotations and the corresponding within-pair variance. This analysis allows us to characterize how increasing annotation depth reveals variability in human judgments, independent of pair selection effects.

\subsubsection{Inter-Annotator Agreement Metrics}

While variance captures distributional heterogeneity in human judgments, we also report standard inter-annotator agreement metrics to quantify consensus beyond chance. We focus on fixed-depth subsets where each pair receives the same number of annotations.

For pairs with exactly two annotations ($n=2$), we compute Cohen's $\kappa$ \cite{cohen1960coefficient}, which measures agreement between two annotators beyond chance:

\[
\kappa_{\text{Cohen}} = \frac{P_o - P_e}{1 - P_e}
\]
where $P_o$ denotes the observed agreement between annotators and $P_e$ represents the expected agreement under chance labeling given the marginal label distributions.

For pairs with exactly three annotations ($n=3$), we compute Fleiss' $\kappa$ \cite{fleiss1971measuring}, a generalization of Cohen's $\kappa$ for multiple annotators:

\[
\kappa_{\text{Fleiss}} = \frac{\bar{P} - P_e}{1 - P_e}
\]
where $\bar{P}$ denotes the mean observed agreement across items and $P_e$ the expected agreement under chance labeling.

Fleiss' $\kappa$ is computed on the subset of pairs with exactly three annotations, which forms the largest multi-annotated group with a fixed number of raters. Subsets with four or more annotations are comparatively small and therefore yield less stable $\kappa$ estimates.

These $\kappa$-based metrics provide a conventional notion of agreement beyond chance, while variance complements them by characterizing the distribution of human responses across annotation depths.

\section{Results}

\subsection{Model–Human Agreement Across Models and Modalities}

\subsubsection{Overall Model–Human Agreement}
Figure 1 summarizes overall model–-human agreement rates aggregated across all task groups. Across text-only models, agreement rates range from 0.57 to 0.69. Among these, TE3L (text-embedding-3-large) exhibits the highest average agreement with human judgments, while other text models cluster more closely around the mid-0.6 range. Multimodal models show consistently lower agreement overall, with rates concentrated between 0.522 and 0.565. Although differences among multimodal models are relatively small, the consistently lower agreement compared to text-only models may suggest that aligning human semantic preferences remains more challenging in multimodal settings. While several models exhibit overlapping intervals, the observed spread in mean agreement indicates systematic differences in alignment behavior across model families rather than random variation alone. These results establish a baseline level of partial alignment between embedding models and human judgments, motivating a more detailed examination of where and how disagreement arises across tasks and modalities.

\begin{figure}[h]
  \centering
  \includegraphics[width=\linewidth]{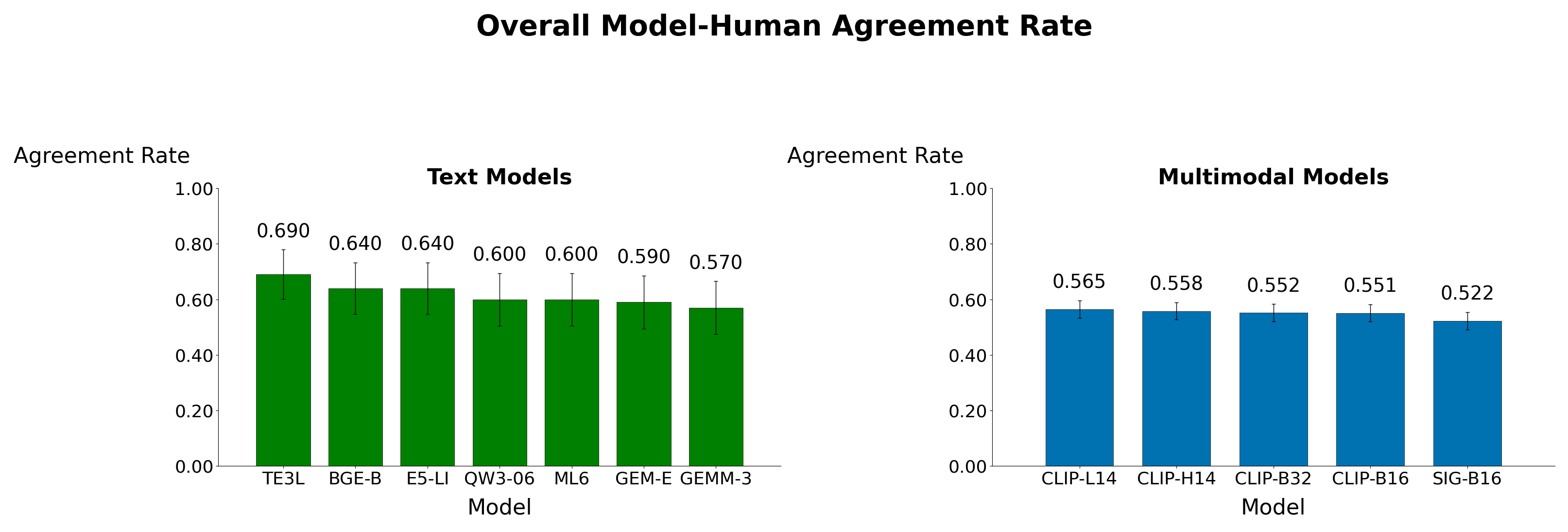}
  \caption{Overall model--human agreement. Mean agreement rates between embedding models and human judgments across evaluated comparisons. Error bars indicate 95\% confidence intervals. Models are evaluated only on tasks within their supported modalities.}
  \Description{Figure 1 is a grouped bar chart showing mean model--human agreement rates for 7 text and 5 multimodal embedding models, with error bars indicating 95\% confidence intervals.}
\end{figure}

\subsubsection{Agreement by Task Group}

Figure 2 breaks down model–human agreement rates for multimodal models by task group, revealing a stable, task-dependent pattern that is not visible in overall averages. Across all evaluated multimodal models, agreement is consistently highest on II comparisons, intermediate on IT comparisons, and lowest on TT comparisons. This ordering holds for every model, indicating that alignment difficulty is systematically tied to comparison type rather than driven by a single model. 

Across models, II agreement rates fall within a relatively narrow band (0.585-0.635), whereas IT agreement is uniformly lower (0.515-0.561) despite being evaluated over a substantially larger set of comparisons (IT: 672 vs. TT/II: 100 each). TT agreement exhibits the lowest and most variable performance across models, with some models falling below 0.50, suggesting frequent divergence from human judgments on purely textual semantic comparisons.

\begin{figure}[h]
  \centering
  \includegraphics[width=\linewidth]{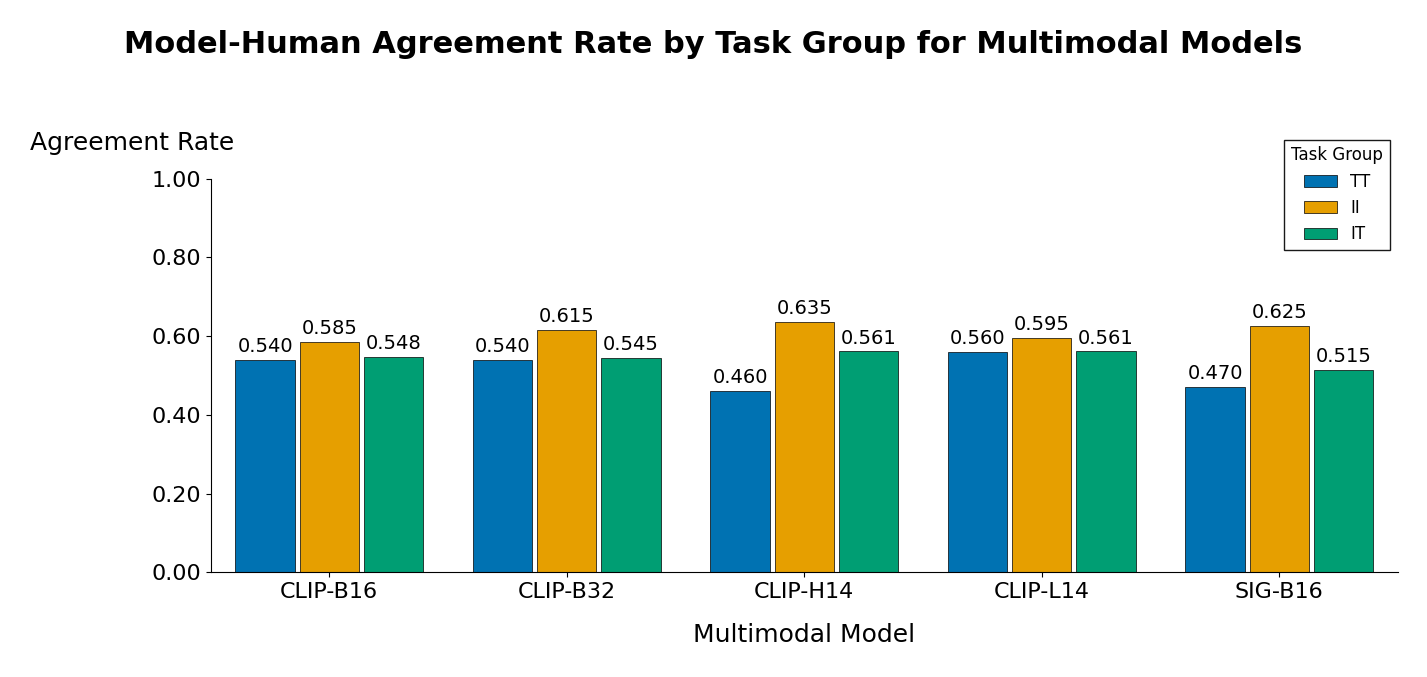}
  \caption{Model–human agreement by task group for multimodal models.
Agreement rates for multimodal embedding models across TT, II, and IT comparison tasks. Agreement is computed as the proportion of comparisons in which the model’s preference matches the human majority choice, with ties scored as 0.5. Reported rates are aggregated within each task group; the number of evaluated comparisons differs by group (TT: 100, II: 100, IT: 672). Models are evaluated only on task groups they support.}
  \Description{Figure 2 is a grouped bar chart showing model--human agreement rates for multiple multimodal embedding models across TT, II, and IT task groups. The horizontal axis lists 5 models, and the vertical axis shows agreement rates. For each model, separate bars correspond to different task groups. Bars represent aggregated agreement proportions.}
\end{figure}

\subsubsection{Inter-Annotator Variability at the Pair Level}
To assess variability in human judgments, we analyze annotation patterns at the level of unique item pairs. Across the cleaned v1 dataset, there are 872 unique pairs, each defined as a specific item pairing presented to participants. For each pair, we compute the proportion of times option B is selected across all annotations for that pair.

Across pairs, the mean proportion of B selections is 0.527, indicating no systematic bias toward either option across the dataset as a whole. However, the variance of B-selection proportions across pairs is 0.180, suggesting substantial variation in the strength of majority preference across item pairs. Some pairs yield strong majority agreement choices, whereas others produce more split judgments, consistent with differences in comparison difficulty or ambiguity.

In contrast to the between-pair analysis above, we examine within-pair variance to assess the consistency of human judgments for the same item pair. We identified 399 pairs (45.8\% of total pairs) with multiple annotations. The mean within-pair variance of 0.1517 across all multi-annotated pairs indicates moderate inter-annotator agreement, suggesting that while annotators generally show consensus, there exists meaningful variability that reflects the inherent subjectivity of semantic similarity judgments.

Our analysis reveals three distinct agreement patterns: 28.8\% of multi-annotated pairs achieved perfect agreement (variance = 0, where all annotators made the same choice), 75.7\% showed majority consensus (where more than 50\% of annotators agreed on the same choice, which includes the 28.8\% with perfect agreement), and 24.3\% resulted in tie cases (variance = 0.25, where exactly half chose A and half chose B). Tie cases occur only for comparisons with even numbers of annotations, since an equal split between options A and B is not possible when the annotation count is odd. Breaking down the majority consensus category, 28.8\% achieved perfect agreement (identical choice) and 46.9\% showed non-perfect majority agreement (where a majority but not all annotators agreed). The presence of tie cases, while representing complete disagreement, is valuable for our evaluation framework as it identifies the genuine diversity in human perspectives on semantic similarity and where different human perspectives are equally valid. The fact that 75.7\% of pairs show majority consensus (with 28.8\% achieving perfect agreement) demonstrates that our dataset captures both clear cases where human judgment converges and uncertain cases where meaningful disagreement exists.

We observed a statistically significant positive correlation between annotation count and within-pair variance (r = 0.27, p < 0.001). Analysis by annotation count groups reveals this trend. Pairs with 2 annotations show a mean variance of 0.1180 and a perfect agreement rate of 52.8\%, with the remaining 76 pairs forming tie cases (161 pairs total). Pairs with 3-5 annotations show a mean variance of 0.1535 and a perfect agreement rate of 28.3\%, with 4 tie cases (53 pairs total). Pairs with 6-10 annotations show a mean variance of 0.1672 and a perfect agreement rate of 13.6\%, with 14 tie cases (110 pairs total). Pairs with 11+ annotations show a mean variance of 0.2000 and a perfect agreement rate of 0\%, with 3 tie cases (75 pairs total). 

To complement the variance-based analysis above, we additionally report standard inter-annotator agreement metrics. For pairs with exactly two annotations, we compute Cohen’s $\kappa$, yielding $\kappa$ = 0.056. For pairs with exactly three annotations, we compute Fleiss’ $\kappa$, yielding $\kappa$ = 0.077. These values indicate modest agreement beyond chance.

This pattern is consistent with the design of our task. Participants are asked to make comparative similarity judgments between two candidate item pairs, and many comparisons may reasonably support more than one interpretation. As a result, even when annotators exhibit meaningful preference patterns, agreement beyond chance may remain moderate.

These $\kappa$ values should be interpreted alongside the variance-based analysis reported above. While $\kappa$ measures agreement beyond chance, within-pair variance captures the structure of human preference distributions across annotators. The moderate variance levels observed across pairs (mean within-pair variance = 0.1517 on multi-annotated items) indicate that annotators often converge on majority preferences while still reflecting diversity in how semantic similarity is interpreted. Because the binary forced-choice task produces nearly balanced label distributions (mean B proportion = 0.527), the expected agreement by chance is relatively high, which in turn leads to lower $\kappa$ values even when majority preferences are present.

\section{Discussion}

\subsection{What Model–Human Agreement Reveals About Embedding Alignment}
These results provide evidence that current embedding models exhibit systematic but incomplete alignment with human semantic judgments, and that the nature of this misalignment depends more strongly on task structure than on model choice within a given modality.

Figure 1 shows that, when aggregated across all task groups, embedding models achieve only moderate agreement with human comparative judgments. Even the strongest-performing text-only model, TE3L, aligns with humans on fewer than 70\% of comparisons, while multimodal models cluster at lower agreement levels, generally below 0.57. Although confidence intervals overlap for several models, the consistent separation between text-only and multimodal model families indicates a systematic gap in alignment behavior rather than random variation.

From a systems perspective, these results suggest that embedding similarity, commonly used in RAG, retrieval, and recommendation pipelines, only partially reflects human semantic preferences. By grounding evaluation in human comparative judgments rather than retrieval accuracy or similarity correlations alone, our benchmark highlights alignment failures that are obscured by conventional embedding evaluations. In practical deployments, these misalignments may surface as subtle but persistent errors in ranking, retrieval, or clustering behavior.

Figure 2 further demonstrates that the task group explains substantially more variance in agreement than model choice among multimodal embeddings. Across all evaluated multimodal models, agreement is consistently highest for II comparisons, intermediate for IT, and lowest for TT. This ordering holds uniformly across models, indicating that alignment difficulty is fundamentally tied to the type of semantic comparison being performed, rather than a particular architecture or training pipeline.

These patterns suggest that multimodal training does not confer uniform benefits across all semantic tasks. Instead, it appears to prioritize visual similarity alignment while leaving cross-modal and textual semantic reasoning comparatively under-aligned. One possible explanation for this pattern lies in the representational trade-offs introduced by multimodal training objectives. Multimodal embedding models are typically optimized to align visual and textual representations within a shared embedding space  \cite{devlin2019bert}. While this objective improves cross-modal correspondence, it may reduce the representational capacity devoted to capturing fine-grained lexical distinctions between words. In contrast, text-only embedding models are trained exclusively on linguistic corpora and therefore specialize in modeling lexical semantic relationships. As a result, multimodal representations may sacrifice some lexical precision in exchange for improved cross-modal alignment. Aggregate agreement metrics obscure this structure, potentially masking failure modes that are highly relevant for real-world systems relying on multimodal embeddings for mixed-media retrieval or reasoning. More broadly, this trade-off highlights a challenge in multimodal representation learning in that optimizing a shared embedding space across modalities may introduce performance differences across semantic comparison tasks.

\subsection{Inter-Annotator Variability and Annotation Depth}
As annotation depth increases, within-pair variance rises while perfect agreement declines. This pattern is expected and desirable, as increasing the number of annotators allows us to capture a broader range of human judgment and more clearly reveal genuine diversity in human perspectives on semantic similarity. 

This trend primarily reflects both statistical constraints (where variance with only 2 annotations is mathematically limited to 0 or 0.25, while more annotations allow intermediate values) and true human judgment diversity (where more annotators reveal genuine variability that may be masked with fewer annotations). Since pairs are randomly generated, pairs with more annotations are simply due to random chance rather than inherent controversy, making this pattern a robust indicator of genuine human judgment diversity rather than a selection effect. The moderate overall variance of 0.1517 (well below the theoretical maximum of 0.25), combined with the majority consensus rate of 75.7\%, demonstrates that annotators show reasonable agreement on most comparisons while also capturing meaningful variability. Standard agreement metrics (Cohen’s $\kappa$ and Fleiss’ $\kappa$) similarly indicate modest agreement beyond chance. This apparent contrast between majority consensus and relatively low $\kappa$ values arises because the binary forced-choice design produces nearly balanced label distributions, which increases the expected agreement under chance. This variability is particularly valuable for evaluating model alignment, as it allows us to assess how well models align with diverse human perspectives rather than just a single “ground truth” label. By including pairs with varying levels of inter-annotator agreement, our evaluation framework can distinguish between models that align with majority human judgment versus those that align with diverse or minority perspectives, providing a more nuanced assessment of human-AI alignment in semantic similarity tasks.

\subsection{Implications for Embedding Evaluation and Downstream Systems}
Our results suggest that embedding alignment should be evaluated with greater sensitivity to both task structure and human preference variability. While inter-annotator analysis shows that human comparative judgments are sufficiently consistent to support evaluation, the presence of meaningful disagreement indicates that alignment cannot be reduced to matching a single, absolute notion of similarity.

A central contribution of Aligned Machine lies in its evaluation design. Unlike conventional embedding benchmarks that treat similarity as a scalar score or retrieval outcome, our dataset evaluates alignment through comparative judgments between two candidate similarity relations. Each evaluation instance asks whether an embedding model ranks one pair of items as more similar than another, mirroring the relative decisions that embeddings support in ranking and selection settings. This design makes it possible to assess whether the ordering induced by an embedding model across comparisons aligns with aggregated human preferences, rather than whether individual similarity scores exceed a fixed threshold.

These findings are particularly relevant for retrieval-driven systems such as RAG, semantic search, and recommendation pipelines, where embeddings are used to rank candidates under the assumption that similarity ordering reflects human relevance. The structured misalignment patterns observed across task types suggest that this assumption holds unevenly, depending on the type of semantic comparison being performed (e.g., TT, II, or IT). Human-centered, comparative evaluation therefore provides a practical diagnostic tool for auditing embedding behavior in systems where ranking decisions interact directly with human judgment.

\subsection{Limitations and Future Works}

The current study is subject to several limitations that we anticipate addressing in future work. First, the dataset is restricted to English-language inputs. While this constraint simplifies analysis and controls for linguistic variation, it limits the generalizability of the findings. Extending the platform to multilingual contexts would enable investigation of how embedding alignment varies across languages and cultural settings, particularly for models trained on heterogeneous corpora.

Second, the current evaluation design focuses on comparative judgments at the concept and pair level and relies on cosine similarity as the primary measure of embedding similarity. While the concept-level comparison task reduces ambiguity and annotation burden, it does not capture more complex compositional or discourse-level similarity judgments at the sentence level. Future iterations could explore hierarchical task designs that bridge concept-level and sentence-level evaluation, allowing finer-grained analysis of alignment across linguistic scales. In addition, our evaluation relies on cosine similarity as the primary measure of embedding similarity. Recent work has questioned the reliability of cosine similarity as a proxy for semantic similarity, noting that embedding space geometry may distort relational interpretation \cite{zhou2022problems, steck2024cosine}. Future work may explore alternative similarity measures or evaluation metrics to better capture semantic relationships in embedding representations.

Third, annotation sparsity and limited comparison coverage remain central limitations in the current dataset. In the v1 dataset, 45.8\% of item pairs received multiple annotations, while the remaining pairs were annotated only once. This limits the precision of within-pair agreement estimates, although the multi-annotated subset still provides sufficient signal to characterize overall agreement patterns. The number of unique comparisons for TT and II remains limited in the current release, which constrains the breadth of semantic relations represented in the benchmark. Because the platform relies on voluntary participation, the collected judgments may also reflect certain sampling characteristics. Prior work has noted limitations of the Moral Machine paradigm, including potential sampling bias and challenges in interpreting large-scale public judgments across cultural contexts \cite{himmelreich2018never}. While the Aligned Machine platform adopts a similar participatory approach to enable large-scale data collection, future iterations may incorporate more systematic participant recruitment and reporting practices to better contextualize large-scale judgment data.
Finally, the current design does not include repeated, identical comparisons across all participants, which prevents direct estimation of inner-annotator consistency. To address this limitation, future versions, such as v2 of the dataset, will ensure a small set of shared anchor comparisons ($n \geq 2$) presented to all participants, enabling computation of inner-annotator agreement and more principled separation of individual noise from genuine judgment diversity. Achieving this goal will require additional iterations of data collection, larger participant pools, and potentially adaptive sampling strategies that prioritize uncertain or high-disagreement pairs.

Future work may explore how HASB can be integrated into training pipelines, active learning schemes, or post-hoc calibration methods.

\section{Conclusion}
This paper introduces Aligned Machine, a human-centered framework for evaluating embedding models through comparative semantic similarity judgments. By focusing on model–human agreement rather than absolute similarity scores, the framework evaluates alignment by directly comparing model-induced preferences with human comparative choices across text, image, and multimodal tasks. This approach reveals systematic differences in how embedding spaces relate to human semantic judgments that are not captured by conventional static benchmarks. Across models and task modalities, we observe consistent variation in model–human agreement, suggesting that embedding alignment depends not only on model architecture but also on the semantic and modal structure of the comparison task. As embeddings increasingly mediate retrieval and recommendation systems, treating human disagreement as signal rather than noise is a necessary step toward more transparent and human-aligned representation learning.

\bibliographystyle{ACM-Reference-Format}
\bibliography{references}

\end{document}